\RequirePackage{fix-cm}
\documentclass[twocolumn,natbib]{svjour3}          % twocolumn
\smartqed  % flush right qed marks, e.g. at end of proof
\usepackage{graphicx}
\usepackage{mathptmx}      % use Times fonts if available on your TeX system
\usepackage[ruled,vlined,linesnumbered]{algorithm2e}
\usepackage[nolist]{acronym}
\usepackage{setspace}
\usepackage{etoolbox}
\usepackage{amsmath}
\usepackage{amssymb}
\usepackage{breakcites}
\usepackage{wrapfig}
\usepackage{xcolor}
\AtBeginEnvironment{algorithm}{\setstretch{1.35}}
\makeatletter
\newcommand{\removelatexerror}{\let\@latex@error\@gobble}
\makeatother

\let\origtau\tau % save the original form of '\tau'
\renewcommand{\tau}{\scalebox{1.44}{$\origtau$}}

\newcommand{\irchi}[2]{\raisebox{\depth}{$#1\chi$}}

\begin{acronym}
\acro{CBRN}{Chemical, Biological, Radiological and Nuclear}
\acro{GDM}{Gas Distribution Mapping}
\acro{GMRF}{Gaussian Markov Random Field}
\acro{UAV}{Unmanned Aerial Vehicle}
\acro{UGV}{Unmanned Ground Vehicle}
\acro{MRO}{Mobile Robot Olfaction}
\acro{GP}{Gaussian plume}
\acro{IP}{isotropic plume}
\acro{CFD}{Computational Fluid Dynamics}
\acro{DSTL}{Defence Science and Technology Laboratory}
\acro{RMSE}{root mean square error}
\acro{SR}{success rate}
\acro{MST}{mean search time}
\acro{SLAM}{simultaneous localisation and mapping}
\acro{IPP}{informative path planning}
\acro{IQR}{interquartile range}
\acro{PDF}{probability density function}
\acro{ATD}{atmospheric transport and diffusion}
\acro{RRT}{rapidly-exploring random trees}
\acro{FMT$^*$}{fast marching trees}
\acro{BIT$^*$}{batch informed trees}
\acro{RGG}{random geometric graph}
\acro{EPSRC}{Engineering and Physical Sciences Research Council}
\acro{DSTL}{Defence Science and Technology Laboratory}
\acro{PSO}{particle swarm optimisation}
\end{acronym}

\begin{document}

\title{Autonomous search of an airborne release in urban environments using informed tree planning

}


\author{Callum Rhodes \and
        Cunjia Liu \and
        Paul Westoby \and
        Wen-Hua Chen%etc.
}


\institute{C. Rhodes \at
              Loughborough University, Loughborough, LE11 3TU, UK \\
              \email{c.rhodes@lboro.ac.uk}           %  \\
           \and
           C. Liu \at
              Loughborough University, Loughborough, LE11 3TU, UK \\
              \email{c.liu5@lboro.ac.uk}
            \and
            P. Westoby \at
            Dstl Porton Down, Salisbury, Wiltshire, SP4 0JQ, UK \\
            \email{pbwestoby@dstl.gov.uk}
            \and
            W-H. Chen \at
              Loughborough University, Loughborough, LE11 3TU, UK \\
              \email{w.chen@lboro.ac.uk}
            }

\date{Received: date / Accepted: date}

\maketitle

%There are \quickcharcount{main} characters.

\begin{abstract}
The use of autonomous vehicles for source localisation is a key enabling tool for disaster response teams to safely and efficiently deal with chemical emergencies. Whilst much work has been performed on source localisation using autonomous systems, most previous works have assumed an open environment or employed simplistic obstacle avoidance, separate from the estimation procedure. In this paper, we explore the coupling of the path planning task for both source term estimation and obstacle avoidance in an adaptive framework. The proposed system intelligently produces potential gas sampling locations that will reliably inform the estimation engine by not sampling in the wake of buildings as frequently. Then a tree search is performed to generate paths toward the estimated source location that traverse around any obstacles and still allow for exploration of potentially superior sampling locations.The proposed informed tree planning algorithm is then tested against the standard Entrotaxis and Entrotaxis-Jump techniques in a series of high fidelity simulations. The proposed system is found to reduce source estimation error far more efficiently than its competitors in a feature rich environment, whilst also exhibiting vastly more consistent and robust results.% These key findings show that an autonomous source search vehicle that can operate successfully in a wide range of environments is achievable when considering path planning around obstacles into the estimation process.
\end{abstract}

\section{Introduction}
\label{intro}
The quick acquisition of accurate estimates of the source of a \ac{CBRN} release is vital in the process of minimising the impact of the resulting hazard and allowing first responders to quickly manage the situation. Doing so with manual sensor probes puts human operators at a high risk of life threatening situations, especially when considering the state of such environments can be highly uncertain. To mitigate this risk, the use of mobile robotic sensors has seen increasing interest as they can be deployed quickly in areas that are inaccessible to humans \citep{Murphy2012}. 

Whilst the use of manually driven \ac{CBRN} robots has already seen use in the field, for example in settings such as nuclear plant decommissioning \citep{Tsitsimpelis2019}, these vehicles must be operated by trained users. If a trained operator is not available or close by at the onset of a disaster event, then the manual nature of the system adds further delay to a process which is heavily time critical. The obvious next step to this problem is to automate the task of data collection, allowing robotic agents to be deployed autonomously.

When considering an autonomous system for a source search task, there are several basic functionalities that the agent needs to possess. Firstly, it should be able to estimate source term parameters as it collects data. This is so that the system can update its belief about the source and use this information to help dictate its next course of action, e.g. collecting data for recursive inference. Secondly, the agent should be able to plan valid trajectories to data sampling locations that help achieve the task of finding the source. This process in feature rich environments is denoted as \ac{IPP}.

Many works, including the pioneering work by \citet{RISTIC20161}, have focused on the first task of source term estimation and only briefly incorporate some form of \ac{IPP} thus having little appreciation to feature rich environments (which cannot be overlooked for real-world \ac{CBRN} incidents). Most of the existing solutions also use a myopic path planner wherein utility of only neighbouring locations are considered (discussed in section \ref{sec:lit_simple} \& \ref{sec:lit_complex}), so may limit the searching efficacy. Therefore, this paper seeks to develop an improved \ac{IPP} solution to the source term estimation problem that is capable of navigating complex environments whilst efficiently carrying out its task of localising an unknown source.

It should be noted that although path planning in complex environments has generated solutions that consider long term trajectories and the overall goal of the agent in a wide array of scenarios, it is not straightforward to bring them into the source term estimation framework. Algorithms that use \ac{RRT} structure are popular in literature due to the flexibility on dynamic constraints, with their spanning trees capable of expanding into any free space. However, directly applying this free spanning feature may not produce good results for source search in an urban environment. In this case, it is intuitive that sampling downwind of buildings is less likely to achieve predictable concentration data than in the open wind direction, due to the influence of buildings on the plume structure (see section \ref{sec:simu} for CFD examples of such a plume structure). Therefore, more samples should be collected in preferential areas compared to obscured areas, so that simple analytic dispersion models can be used to interpret the measurements. This principle may be violated by the one-sample-per-batch approach of \ac{RRT} as branches are grown heuristically towards a sampled state and therefore branches can be grown into undesirable regions, regardless of the desired sampling frequency of said region. Recent works on \Ac{FMT$^*$} \citep{Janson2015} and \ac{BIT$^*$} \citep{Gammell2015} are two sampling based approaches that sample the whole environment in a single batch to create a \ac{RGG} in which spanning trees are expanded. Such multi-sample batch methods allow a custom sampling distribution to be enforced so that undesirable regions are sampled with a lower frequency and therefore a variant of this technique will be explored in the proposed informed tree approach. By employing a sampling based path planner, the system can free itself of deterministic path planning choices and perform adaptively across varying environment scales and complexity. 

Another requirement to be considered is the efficiency of the search, which requires a good balance between exploration and exploitation. The goal location in a conventional path planner is specified in some way, so that it can be directly exploited. For example, in \citep{Gammell2015} the use of a goal set is postulated alongside the single goal state. When combining with a Bayesian inference framework to estimate the source location (i.e. navigation goal), there is not a single (or set) location, but instead it is described by a \ac{PDF}. To this end, inspired by the Dual Control principle introduced in \citep{Chen2020}, the goal state used by the proposed informed tree search algorithm is modified, so that the tree can be spanned iteratively towards the source while accounting for the uncertainty of the source term estimation,

By addressing the above technical challenges, this paper develops a more powerful and more applicable IPP framework for searching an unknown \ac{CBRN} source in an urban environment. This paper is organised as follows. In Section \ref{sec:literature}, relevant works are reviewed to justify the novelty of this work. Section \ref{sec:problem} formulates the problem to be considered, followed by technical solutions in Sections \ref{sec:dist} - \ref{sec:utility}. The proposed algorithm is tested and verified in Section \ref{sec:simu} using a high fidelity dataset and the conclusions are provided in section \ref{sec:conc}.

\section{State of the art} \label{sec:literature}
\ac{CBRN} related robotics has seen a swell of research interest in recent years, as recently summarised in \citep{Monroy2019}, due to the ever increasing computational capabilities of small onboard chipsets and chemical sensors that can be easily fitted to mobile platforms including small UAVs. Coupled with their mobility to collect large amounts of data at any location, these systems are highly beneficial compared to the traditional approach of sparse static sensors running alongside complex CFD models that can take several days to resolve. To enable these mobile systems for source localisation tasks, both online estimation and motion planning functions need to be developed. 

%To leverage their mobility, computationally lightweight data driven estimation algorithms have been favoured for use on mobile robots.

\subsection{Estimation}
Probabilistic estimation algorithms can be split into two categories, i.e., using parametric models and non-parametric models. In parametric estimation algorithms, the \ac{PDF} of the underlying parameters of the \ac{ATD} model will be established. Examples of light-weight \ac{ATD} models that have been applied to mobile robots include the \ac{GP} model \citep{Wang2018} and the \ac{IP} model \citep{Vergassola2007}. These models describe the expected concentration at a given location under defined source terms and environmental conditions. These simple models have drawbacks in that they make strong assumptions about the source (such as a single source, constant release and uniform wind fields), however their computational efficiency lends towards the inclusion in probabilistic frameworks. 

Non-parametric models for gas localisation include Gaussian Process \citep{Hutchinson2019}, Kernel DM+V \citep{Lilienthal2009} and Gaussian Markov random fields \citep{G.Monroy2016}. The number of parameters in these models is not fixed and therefore less assumptions are made about the gas distributions. To account for the transportation of particles, GMRF and Kernel DM have further additions to account for wind direction in GW-GMRF \citep{Gongora2020} and Kernel DM+V/W \citep{Asadi2017}. This leads to the ability to account for multi modal distributions and the inclusion of obstacles in the environment. A major drawback of these methods are that they are poor at estimating outside of sample locations and require a large and varied set of data to accurately estimate the distribution. For a source term estimation case, this leads to the case that if the area near the source location itself cannot be sampled (such may be the case in an urban environment), then non-parametric models tend to be unable to accurately estimate the source. Furthermore, these models tend to be computationally expensive to iteratively calculate. Note that the framework proposed in this paper can use both model types (e.g. \citep{Rhodes2020}), therefore leaving flexibility in the system. However, a parametric model is used in this work, not only due to its computational efficiency and wide spread adoption in the literature, but also because one of the motivations of this work is to show that such simple \ac{ATD} models are adequate to inform robotic source localisation in urban environments, given features can be accommodated in the path planning algorithm. 

\subsection{Source search in simple environments}
\label{sec:lit_simple}
Motion planning of mobile robots plays a key role in many IPP frameworks for environment monitoring. However, in the literature, motion planning for source search is generally limited to goal selection i.e., which place should be sampled next. To solve the goal selection problem that is inherent in an autonomous system, there are three classes of algorithms employed: coverage based, bio-inspired and information theoretic. Coverage based planners (e.g. \citep{Hombal2010,Galceran2013}) rely on predetermined trajectories to maximise coverage of the search area in a systematic manner. These algorithms are incredibly efficient and easy to implement but are decoupled from the estimation side of the system. Therefore, they can be ineffective and difficult to scale. Bio-inspired methods such as Anemotaxis \citep{Harvey2008} and Chemotaxis \citep{Dhariwal2004,Russell2003} use instantaneous concentration and anemometry measurements to guide robots based on the local concentration gradients. These methods are computationally lightweight and are often employed for use in swarm robotics \citep{Marjovi2014,Wisnu2007} where resources are limited. However, they are heavily reliant on the presence of data and do not perform well in large scale scenarios or with sparse measurements, since they only consider immediate reward in their locality.

The third class, and the method that is leveraged in the proposed system, is the information theoretic approach. Information theoretic approaches exploit the belief of the system state and try to take actions that reduce the uncertainty of a given estimate. Given this property, information theoretic approaches are inherently coupled to the estimation process and require some form of metric to quantify uncertainty. Within source term estimation, both Infotaxis \citep{Vergassola2007} and Entrotaxis \citep{Hutchinson2018} have been successfully employed for sparse search tasks. Infotaxis is concerned with reducing entropy based on the expectation of the posterior distribution, whereas Entrotaxis considers the entropy reduction based on the predictive measurement distribution. In a comparison between bio-inspired searches and information theoretic searches \citep{Voges2014}, the information theoretic solution is found to be more effective in problems which exhibit sparse data, thus performs well in real-world experiments \citep{Hutchinson2019b, Hutchinson2019c}. Sparse measuring conditions are more conducive towards urban environments since complex geometry can obscure the plume from much of the search domain. %A hybrid information theoretic approach is taken in \citep{Rhodes2020} in conjunction with a non-parametric gas distribution model and is also found to be superior to a Chemotaxis based solution in urban environments. 
%To this end, an information theoretic approach is taken, specifically using Entrotaxis, as the metric for calculating the utility of future sampling actions.

It is noted that the vast majority of research items that focus on the motion planning aspect of source search do so in an open environment and therefore this gap between open and urban scenarios is the key motivation for the research presented.

\subsection{Source search in complex environments}
\label{sec:lit_complex}
Path planning in complex environments for source search has some studies but many do so in a heavily constrained environments and therefore are not optimised for the challenges of real urban scenarios. In \citep{Marjovi2011}, multi robot mapping and source localisation is performed using an anemotaxis approach wherein \ac{SLAM} is performed until a threshold concentration is found, upon which the robot switches to an anemotaxis search. As with all gradient-based approaches, this system requires an increasing number of agents to cope with large scale situations. In \citep{Zou2014}, a \ac{PSO} method is proposed that accounts for obstacles that each sensing agent may encounter by proposing new directions that do not intersect with the obstacles. This simplistic approach to obstacle avoidance has clear success with swarm implementations, however, to enable a single agent to efficiently cover a large area within a time budget, we argue more advanced methods are needed that plan adaptively in the longer term as opposed to maximising within a deterministic set of neighbouring points.

With the informed tree search algorithm we seek to address the scaling problem with adaptive sampling of the environment so that the system is broadly independent of the scale. This also means that a multi agent approach is not required to attain positive results. However, it is appreciated that multi agent approaches are generally more time efficient than their single agent counterpart (at the expense of increased resources).

In model-based search techniques, \citep{Khodayi-Mehr2019} propose a solution that uses an \ac{ATD} model solved via partial differential equations to identify a source. This work uses the Fischer Information matrix of the source parameters to select a sequence of future waypoints and is shown to be capable of operating in small non-convex domains but is not proven for larger urban scenarios.

In \citep{Zhao2020}, the Entrotaxis-jump algorithm is proposed for source search in a large-scale road network. Entrotaxis-jump combines Entrotaxis with an intermittent search strategy that allows a myopic agent to traverse around obstacles if the utility of sampling in the direction of the obstacle is high. Whilst this method successfully increases the performance of Entrotaxis in urban environments, it assumes a simple Gaussian-like dispersion model adequately reflects the dispersion characteristics of a source release in a dense urban environment. Moreover, the trajectory generation is not explored in the search process and therefore for more complex geometries (i.e. not a road network), it is unknown if this method will be able to successfully navigate towards a source. Nevertheless, based on the findings of \citet{Zhao2020} it is clear that there is a benefit to intelligent goal selection methods in urban scenarios compared to those in classical source search motion planning.

Further to this work, Zhao et al. also propose a searching method based on Entrotaxis for escaping forbidden zones in a source search scenario \citep{zhao2020b}. This paper proposes a planning technique to avoid the sampling agent becoming trapped in its locality based on the degrees of free travel around a location. This technique is shown to be effective in a block discretised scenario and is not shown how it could be applied to more complicated geometric scenarios (such as the urban case). Contrary to this, the BIT* path planning technique can also be used to plan escape routes around obstacles and can do so around non-block structures. 

The information theoretic approach has also been seen in other applications involving complex environments. For example, \citet{Schmid2020} developed a sampling based path planner using \ac{RRT} to evaluate the utility of sampling the environment at a specified location along a trajectory. This method allowed a sampling agent the ability to determine efficient sampling locations whilst navigating in complex environments, inherently coupling these two aspects. However, the nature of the work is to explore the unknown environment, rather than locating the release source.

Recently, An et al. present an urban source search algorithm namely receding-horizon RRT-Infotaxis \citep{an2022receding}. This work leverages the standard RRT path planning technique along with the Infotaxis method of determining future sample location utility (similar in framework to our \ac{BIT$^*$} and Entrotaxis technique). In addition, this method also attempts to predict the sampling utility along a multi-step trajectory in a receding-horizon fashion by summing potentials. Whilst predicting utility multiple steps into the future may offer more efficient movement choices, due to the formulation of the utility calculation, this demands a fixed step size between consecutive samples (to cancel out traversal cost considerations) which violates our chosen batch sampling planner. We opt in favour of more robust obstacle avoidance and the capability to incorporate custom sampling distributions.

By employing more advanced path planning techniques that have been adopted in other fields and customising these methods to suit the challenges of the source search problem, we aim to bring a new level of operational efficiency and robustness to the field of \ac{CBRN} related robotics.

\subsection{Contributions}

Based on the literature review, we present the contributions of this paper that address the issues raised above.

{Our first contribution is to show that simple atmospheric dispersion models (such as the \ac{IP} model) can be used to guide a robot to localise a release source in a complicated urban environment.} To aid in this task, we propose a novel sampling distribution that identifies regions in the wake of buildings where a large model discrepancy between the modelled plume and the actual flow would be seen. By sampling in these regions with a lower frequency, we can collect samples that will more reliably inform the inference engine of the true source. This also shows how preferential sampling can help make the search process more efficient and leaves the door open for future work investigating other informed distributions under the same planning framework.

The second major contribution of this paper is the proposed informed tree search algorithm. Based on the \ac{BIT$^*$} concept, the novel tree search method creates branches that extend toward the expected source location (increasing convergence speed) whilst navigating through areas in the wake of buildings with a lower frequency. The informed tree is then either pruned or blossoms to meet the computational requirements of the information utility function, providing obstacle free informative trajectories for the autonomous agent to follow. This second contribution moves away from previous works that rely on sampling at deterministic future locations (such as $\uparrow, \rightarrow, \downarrow, \leftarrow$) and introduces an adaptive sampling framework that balances the trade-off between exploration and exploitation in desirable regions.

\section{Problem statement} \label{sec:problem}
The source search problem is formulated under the \ac{IPP} framework in this Section. 

Let $\mathbf{X} \subset \mathbb{R}^2$ be the state space of the search and planning problem, $\mathbf{X}_{\mathrm{obs}} \subset \mathbf{X}$ be the states in collision with obstacles. Thus, the set of admissible states can be expressed as $\mathbf{X}_{\mathrm{free}} := \mathbf{X} \setminus \mathbf{X}_{\mathrm{obs}}$. Let $\mathbf{s} \in \mathbf{X} $ be the source location, $\mathbf{x}_{k} \in \mathbf{X}_{\mathrm{free}}$ be the robot position at sampling instant $k$ and $\mathbf{X}_{\mathrm{goal}} \subset \mathbf{X}_{\mathrm{free}}$ be the set of goal region. A collision-free path is continuous mapping $\sigma: \mathbb{R} \xrightarrow{} \mathbf{X}_{\mathrm{free}}$. Specifically, we define $\sigma_{i}^{j}(s)$, $ s \in [0, \,1]$, a path from $\sigma_{i}^{j}(0) = \mathbf{x}_{i}$ to $\sigma_{i}^{j}(1) = \mathbf{x}_{j}$. The traversal length of the path is denoted as $c(\sigma_{i}^{j})$.

The problem considered in this work is to guide the robots to explore the free space to find the source location $\mathbf{s}$ and eventually navigate to the goal region inclusive of the source location, such that $\mathbf{x}_{k} \in \mathbf{X}_{\mathrm{goal}}(\mathbf{s})$. However, directly finding an optimal path or feasible path from initial position $\mathbf{x}_{init}$ to $\mathbf{s}$ is not possible, since the source location is unknown to the robot. In this case,  a recursive \ac{IPP} framework will be structured to address this problem.  

\begin{problem}[source term estimation]
At each sampling time $k$, the robot takes a measurement of the local chemical concentration $z_{k}(\mathbf{x}_{k})$, which in conjunction with historical readings $\mathcal{Z}_{k} = \{ z_{k}(\mathbf{x}_{k}), \mathcal{Z}_{k-1} \}$, can be used to estimate the source term $\Theta$, in the form of its posterior distribution, i.e., $p(\Theta|\mathcal{Z}_k)$.
\end{problem}
 
{Note that the source term $\Theta$ normally consists of source location $\mathbf{s}$, release rate $Q$ and other relevant parameters that can be used to characterise an airborne release. In this study we utilise the \ac{IP} model, the parametrisation of which is shown in Eq. \eqref{eqn:IP_param} (see \citep{Hutchinson2019b} for more details).  
\begin{align} \label{eqn:IP_param}
    \Theta_k=\big[\mathbf{s}^{T} \mkern9mu Q \mkern9mu u \mkern9mu \phi \mkern9mu d \mkern9mu \origtau \big]^{T} 
\end{align}
where $Q$ is the release rate of the source (g/s), $u$ is the wind field speed (m/s) with direction $\phi$ (deg), $d$ is the diffusivity of the hazard in air (m$^2$/s) and $\origtau$ is the average lifetime of the emitted particle (s). Using this model, for a given $\Theta_k$, the expected concentration that a sensor will record at position $\mathbf{x}_k$ is calculated using:
\begin{multline}
    C(\mathbf{x}_k|\Theta_k)=\frac{Q}{4\pi d\|\mathbf{x}_k-\mathbf{s}\|_2}\exp\bigg[\frac{-\|\mathbf{x}_k-\mathbf{s}\|_2}{\lambda}\bigg] \\
    \times \exp\bigg[\frac{-x_k-x_s u\cos \phi}{2d}\bigg]
    \times \exp\bigg[\frac{-y_k-y_s u\sin \phi}{2d}\bigg] \\
    \label{eqn:IPmodel}
\end{multline}
where, $\lambda=\sqrt{\frac{d\origtau}{1+(u^2\origtau)/(4d)}}$.
} Note that to facilitate the discussion we use $\theta_x^{(k)}$ to denote the estimated source location at sampling instant $k$.
\begin{problem}[\ac{IPP}]
Let $\sigma_{k}^{k+1}$ be a collision-free path that can be executed by the robot, starting from the robot's current location $\mathbf{x}_{k}$ to an end location $\mathbf{x}_{k+1}$. Let $\Sigma$ be the set of such non-trivial paths to be constructed. The IPP problem is then formally defined as the search for a path, $\sigma^{*} \in \Sigma$, that minimise a utility function $\Psi(\cdot)$, such that
\begin{equation}
    \sigma^{*} := \arg \min_{\sigma \in \Sigma} \{ \Psi(\sigma) | \sigma(0) = \mathbf{x}_{k}, \, c(\sigma) \leq \bar{c}  \}\\
    \label{eqn:IPP}
\end{equation}
where $\bar{c}$ is the upper bound of the path length.
\end{problem}
Note that the objective of the path planning problem is to find the most informative sampling location at the end of the path $\sigma^{*}$. This is because the chemical sensing robot normally takes point measurements to accommodate the response time of the chemical sensor. 
%This also means there may not be a specific goal location or region in the path planning formulation. 

The proposed framework is to recursively solve Problem 1 and 2 such that the robot can be guided to the source region $\mathbf{X}_{\mathrm{goal}}(\mathbf{s})$. At each sampling instant $k$, the robot takes the sensor reading $z_{k}$ to update the source term estimation $p(\Theta|\mathcal{Z}_k)$. Such a posterior distribution can be used to inform the design of the set $\Sigma$, so that Problem 2 can be constructed and subsequently solved to generate the next sampling location at the end of $\sigma^{*}_{k}$. 

\begin{figure}[!t]
    \centering
    \includegraphics[width=\columnwidth]{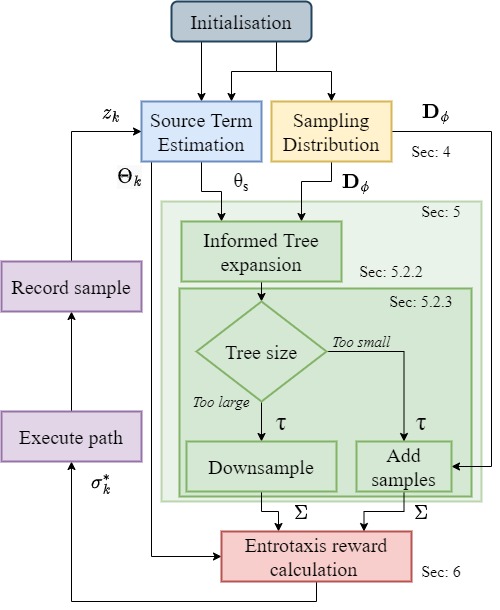}
    \caption{Architecture of the proposed source search algorithm}
    \label{fig:systemOverview}
\end{figure}

In this work, Problem 1 is solved by using an established particle filter developed in \citep{Hutchinson2019b}, so its implementation detail is skipped for the sake of brevity. The key challenge that remains open is how to efficiently construct the set of candidate paths $\Sigma$ in Problem 2, which should 1) reduce the chance of taking samples downwind of buildings, 2) guarantee collision-free control actions and 3) strike a good balance between exploitation and exploration. To this end, an informed tree search algorithm is developed and integrated into the proposed framework as outlined in Fig. \ref{fig:systemOverview}.  Each part of the downward running system will be explained in further detail in the order that they are performed during one iteration of the planning loop. 

\section{Generation of sampling locations}
\label{sec:dist}
Many STE algorithms for predicting the source parameter distribution (e.g. \citet{RISTIC20161,Hutchinson2018,Zhao2020}), whilst proven in ideal open environments, have drawbacks when used in a feature rich environments. Because obstacle interactions with the gas dispersion are not accounted for (and the use of such a computationally expensive model is unsuitable for mobile robotics), the intelligent use of how sample locations are chosen can be leveraged to mitigate the incongruity of the model and the physical system.    

As defined in the problem statement, the set of admissible states that are considered for the search and planning problem is defined as $\mathbf{X}_{\mathrm{free}}$. When using a sampling based path planner, traditionally a single sample state $\mathbf{x}_m$ is drawn such that $\mathbf{x}_m \gets \{\mathbf{D}\sim\mathbf{U}(\mathbf{X}_\mathrm{free})\}$, where $\mathbf{D}$ represents the uniform distribution of states that exist within free space. We also define the notation $\mathbf{X}_m \gets \{\mathbf{D}\sim\mathbf{U}(\mathbf{X}_{\mathrm{free}})\}_{1:N}$, where $\mathbf{X}_m$ is a set of samples with size $N$, uniformly drawn from free space.

In general traversability planning, the definition of $\mathbf{X}_\mathrm{free}$ is adequate for dictating how sample states should be drawn and therefore a uniform distribution is most often used. However, as stated in \citep{Karaman2011}, the sampling framework can extend to any distribution with a density bounded away from $0$ upon $\mathbf{X}_\mathrm{free}$. In source term estimation, (as discussed in the introduction) the robot should sample in areas where it is most likely to interact with the target source and also in areas which are more likely to accurately predict the source. When obstacles are present in the flow field, this is not uniform over the free space since there is a modelling discrepancy between the \ac{IP} model and the actual flow. 

Obstacle interactions with scalar wind fields lead to complex flow dynamics that take significant resources to resolve. However, it is clear that a particle in a laminar wind field will generally move in the wind direction $\phi$, unless obstructed by an obstacle. Obstacles create isolated areas (wake) behind the obstacle that disrupt flow and create a disparity between what the model predicts and the real flow. It is in these areas that is less likely for a robot to sample the source plume predictably since the estimation model implemented does not account for obstacle interactions (due to computational constraints). {Therefore, a sample distribution $\mathbf{D}_\phi$ should be attained that stipulates the robot to sample less in these areas that are likely to observe contradictory measurements.}

$\mathbf{D}_\phi$ is derived (similarly to \citep{Bellingham2002}) by calculating the divergent effect that an obstacle would have on a particle entering the search space using Dijkstra's search (detailed below and shown in Fig. \ref{fig:dijkstra}). The sample generation technique, whilst being significantly quicker to compute than CFD modelling, can be expensive to calculate for large maps and therefore should be performed at a reasonable resolution. Samples can be drawn repeatedly from the same distribution assuming the conditions that the wind direction $\phi$ does not change significantly and that the obstacle map is static. The sample generation methodology comprises one of the main new contributions to the field of source search and feeds directly into the second new contribution, the informed tree search. It should be noted that this method is adopted for its ability so generate $\mathbf{D}_\phi$ quickly and with little a-priori environmental information. Given an infinite budget, similar approaches using more complex modelling could also be implemented which may improve the search efficiency further. Due to the design choice of implementing the BIT$^*$ method, any informed distribution for preferential sampling may be exploited.

\subsection{Sample distribution algorithm}

\begin{figure}[!t]
    \centering
    \includegraphics[width=0.9\columnwidth]{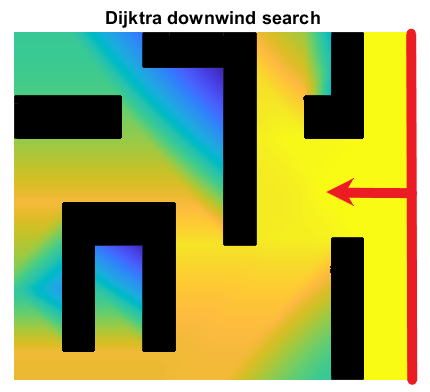}
    \caption{Sample distribution of the likely trajectory of an inlet gas particle, attained from Dijkstra's search on an occupancy grid. Yellow indicates a high likelihood whilst blue indicates a low likelihood. Red arrow shows the wind direction.}
    \label{fig:dijkstra}
\end{figure}

To generate a probability distribution that reflects the obstacle interactions with the wind field, firstly a set of starting states that represents the inlet of a particulate to the search space $\mathbf{X}$ is defined as $\mathbf{X}_{\mathrm{inlet}}$ (red line in Fig. \ref{fig:dijkstra}). These inlet states are akin to the inlet condition of a CFD model thus the location of these states is dependent on the wind field direction $\phi$, which can be initiated with the prior of $\mathbb{E}(\theta_\phi)$ from the Bayesian inference. Dijkstra's search is then performed on the discrete state obstacle map (e.g. an occupancy grid) using $\mathbf{X}_{\mathrm{inlet}}$ as starting conditions. The vertices of the Dijkstra network $\mathbf{V}_{\mathrm{obs}}$ are defined as all $\mathbf{x} \in \mathbf{X}_{\mathrm{free}}$ and edges of traversal $\mathbf{E}_{\mathrm{obs}}$ lie between obstacle free adjacent vertices.  This then generates the average cost $\mathbf{C}_{\mathrm{obs}}$ of getting from all $\mathbf{x} \in \mathbf{X}_{\mathrm{inlet}}$ to all $\mathbf{x} \in \mathbf{X}_{\mathrm{free}}$. A second cost map, $\mathbf{C}_{\mathrm{open}}$, is also calculated using Dijkstra's search on the obstacle free map using the same $\mathbf{X}_{\mathrm{inlet}}$ condition. The vertices of the second network $\mathbf{V}_{\mathrm{open}}$ are defined as all $\mathbf{x} \in \mathbf{X}$ and edges of traversal $\mathbf{E}_{\mathrm{open}}$ lie between any adjacent vertices. This second cost map represents how an inlet particle would traverse the domain uninterrupted by the obstacles. $\mathbf{C}_{\mathrm{open}}$ is then subtracted from $\mathbf{C}_{\mathrm{obs}}$ leaving a final cost map $\mathbf{C}_\phi$ that represents how the obstacles have negatively interrupted the wind field. $\mathbf{C}_\phi$ can then be used to give the probability distribution $\mathbf{D}_\phi$, by adding the minimum value of $\mathbf{C}_\phi$. $\mathbf{D}_\phi$ is now bound away from 0 upon $\mathbf{X}_{\mathrm{free}}$ and can be used to draw samples from which to grow the informed tree. This process is summarised in Algorithm \ref{alg:sampleGeneration}.

\begin{figure}[!t]
\removelatexerror
\begin{algorithm}[H]
    \label{alg:sampleGeneration}
    \SetAlgoLined
    \SetKwFunction{proc}{init}
    \caption{Sample Generation ($\mathbf{X},\mathbf{X}_{inlet}$)}
    \SetKwProg{myproc}{Procedure}{}{}
    \myproc{\proc{}}{
    $\mathbf{V}_{\mathrm{obs}} \gets \mathbf{x} \in \mathbf{X}_{free}$ \\
    $\mathbf{C}_{\mathrm{obs}} \gets \sum\text{Dijkstra}(\mathbf{V}_{\mathrm{obs}},\mathbf{E}_{\mathrm{obs}},\forall \mathbf{x} \in \mathbf{X}_{inlet})$ \\
    $\mathbf{V}_{\mathrm{open}} \gets \mathbf{x} \in \mathbf{X}$ \\
    $\mathbf{C}_{\mathrm{open}} \gets \sum\text{Dijkstra}(\mathbf{V}_{\mathrm{open}},\mathbf{E}_{\mathrm{open}},\forall \mathbf{x} \in \mathbf{X}_{inlet})$ \\
    $\mathbf{C}_{\phi} \gets \mathbf{C}_{\mathrm{open}}-\mathbf{C}_{\mathrm{obs}}$ \\
    $\mathbf{D}_{\phi} \gets \mathbf{C}_{\phi}+\arg \displaystyle \min_{x \in X_{free}}\mathbf{C}_\phi(\mathbf{x})$ \\
    }
    \SetKwFunction{proc}{sample}
  \SetKwProg{myproc}{Procedure}{}{}
    \myproc{\proc{$N$}}{
    $\mathbf{X}_m \gets \{\mathbf{D}_\phi(\mathbf{X}_{free})\}_{1:N}$\\
    \KwRet $\mkern9mu \mathbf{X}_m$
    }
    \end{algorithm}

\end{figure}

\section{IPP Informed tree search}

\begin{figure*}[!t]
    \centering
    \includegraphics[width=\textwidth]{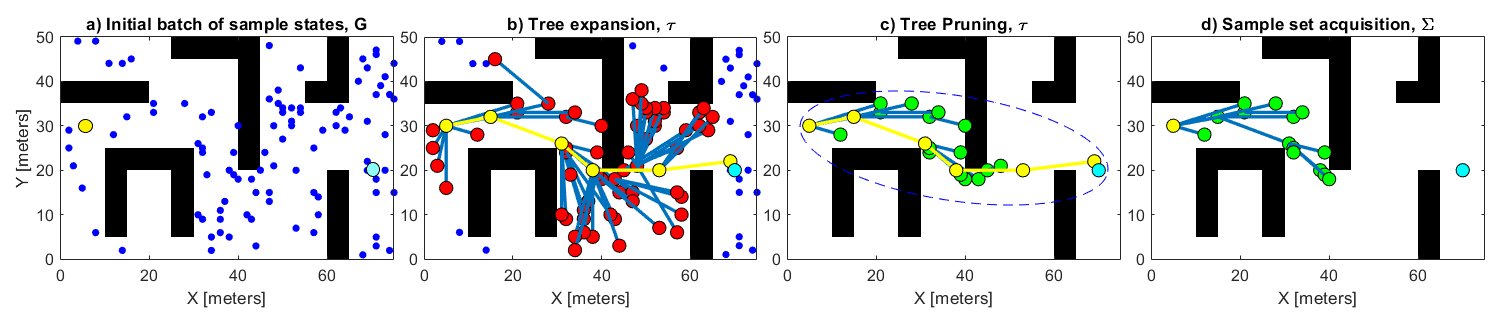}
    \caption{Informed tree search with a starting location $\mathbf{x}_k=[5,30]$ (yellow circle) and goal location $\mathbf{x}_{goal}=[70,20]$ (cyan circle). a) The initial batch of sample states $\mathbf{x} \in \mathbf{G}$ (blue dots) drawn from the discrete probability distribution $\mathbf{D}_\phi$ where $N=100$ and $\theta_\phi$ is in the negative x-direction. b) Tree expansion procedure until the goal is found, tree vertices $\mathbf{v} \in \mathbf{V}$ shown with red circles, tree edges $\mathbf{(v,w)} \in \mathbf{E}$ shown with blue lines and the shortest path (of length $\mathbf{c}_{best}$) to $\mathbf{x}_{goal}$ shown by yellow edges and circles. c) Tree pruning with remaining vertices with green circles and the ellipse that represents the pruning criterion shown with a blue dashed line. d) Downsampling of the pruned tree to contain the first $|\Sigma|=15$ vertices in the tree.}    
    \label{fig:downsample}
\end{figure*}

\begin{figure*}[!t]
    \centering
    \includegraphics[width=\textwidth]{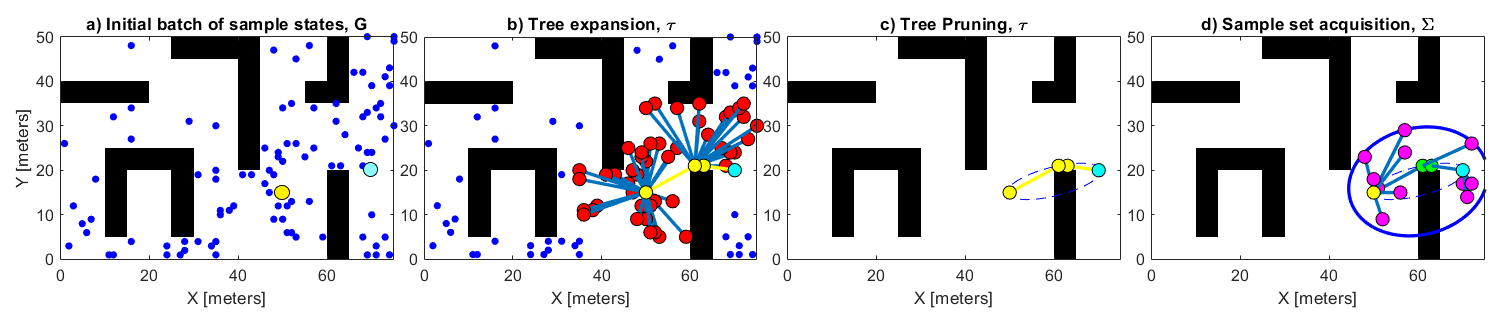}
    \caption{Informed tree search with a starting location $\mathbf{x}_k$ [50,15] (yellow circle) and goal location $\mathbf{x}_{goal}$ [70,20] (cyan circle). a) The initial batch of sample states $\mathbf{x} \in \mathbf{G}$ (blue dots) drawn from the discrete probability distribution $\mathbf{D}_\phi$ where $N=100$ and $\theta_\phi$ is in the negative x-direction. b) Tree expansion procedure until the goal is found, tree vertices $\mathbf{v} \in \mathbf{V}$ shown with red circles, tree edges $\mathbf{(v,w)} \in \mathbf{E}$ shown with blue lines and the shortest path (of length $\mathbf{c}_{best}$) to $\mathbf{x}_{goal}$ shown by yellow edges and circles. c) Tree pruning with remaining vertices with green circles and the ellipse that represents the pruning criterion shown with a blue dashed line. d) Adding of further samples within the expanded ellipse (dark blue solid line) given a source position uncertainty $\sigma_{\theta_x}=5m$. New samples added are shown with magenta circles.}
    \label{fig:addsample}
\end{figure*}

The second main contribution of the paper is the informed tree procedure, which acquires a set of obstacle free trajectories that help the robot achieve its goal of localising an unknown source within a feature rich search space. The informed tree search procedure can be separated into two distinct parts: tree expansion and path selection. Tree expansion is concerned with finding admissible paths to the goal location whereas path selection decides the set of paths to be evaluated by the utility function. Both parts are described below and further detail is given in section \ref{sec:tree_alg}.

To establish possible gas sampling locations and an initial set of admissible paths, an RGG is defined which contains a set of sampled states, $\mathbf{G} \in \mathbf{X}_{\mathrm{free}}$ (Fig \ref{fig:downsample}a, \ref{fig:addsample}a). Samples are drawn in weighted free space, of sample size $N$, with the weighting on samples derived from the likely gas particle path distribution $\mathbf{D}_\phi$ described in Section \ref{sec:dist}. The parameter $N$ is chosen such that the informed tree search can fully explore the area $\mathbf{X}_{\mathrm{free}}$. The base of the tree is set to $\mathbf{x}_k$, and the goal set is defined as the $k_n$-nearest $\mathbf{x} \in \mathbf{G}$ to the best estimate of the source location (other definitions of goal sets are also applicable). For the rest of the paper, $\mathbb{E}(\mathbf{s})$ is used as the best estimate. 

The tree is then heuristically constructed towards the goal set, in a similar manner to \citep{Gammell2015}, until a valid path is found (Fig \ref{fig:downsample}b, \ref{fig:addsample}b). However unlike \citep{Gammell2015}, after one batch search has been completed, another batch is not initiated. This is because we are not trying to acquire the shortest trajectory towards the estimated source location. Instead, we need only a set of admissible and exploratory paths to the source that can guide the search in a way that will converge on the source location. 

Once a path to the goal set is found, the tree is pruned (Fig \ref{fig:downsample}c, \ref{fig:addsample}c), and then must be either upsampled or down-sampled in order to match the computational requirements of the \ac{IPP}. This is driven by the desired number of paths $|\Sigma|$, to be calculated in \eqref{eqn:IPP}. If the tree contains paths $\geq |\Sigma|$, then the pruned tree is downsampled, as shown in Fig \ref{fig:downsample}d, to the first $|\Sigma|$ paths stemming from $\mathbf{x}_k$. If more samples are required, then the tree is expanded by taking further samples (again from weighted free space) that extend the tree as shown in Fig \ref{fig:addsample}d. This expansion is bound within an ellipse that accounts for the uncertainty of the source estimate in $\theta_x$ (derived in section \ref{sec:tree_alg_acq}). Once the path set $\Sigma$ has been established, the set is passed for utility calculation. {This process is then repeated each time a concentration measurement updates the particle filter as $\mathbb{E}(\mathbf{s})$, and therefore the goal set, will change after each update (since  $\mathbb{E}(\mathbf{s})$ is given as weighted expectation). The previous sampled set of nodes in the network, $\mathbf{G}$, can be recycled for efficiency or resampled for diversity.} The detailed derivation of informed tree search is provided in this section.

\subsection{Notation}
To facilitate the derivation of the informed tree search algorithm, some notations are defined in this sub-section. The function $\mathbf{\widehat{g}(x)}$ relates to the estimated cost to come from the start (i.e., $\mathbf{x}_{k}$) to a state $\mathbf{x} \in \mathbf{G}$, and $\mathbf{\widehat{h}(x)}$ is the estimated cost to go from the state to the goal set $\mathbf{X}_{goal}$. These functions can be evaluated using the Euclidean distance between two states. Then, $\mathbf{\widehat{f}(x)}$ is the estimated cost from the start $\mathbf{x}_k$, to the goal set,  given the path passes through $\mathbf{x}$, i.e. $\mathbf{\widehat{f}(x)}:=\mathbf{\widehat{g}(x)}+\mathbf{\widehat{h}(x)}$. If the current best solution to the goal set is defined as $\mathbf{c}_{best}$, then this function can define a subset of states that potentially give a better solution to the goal i.e. $\mathbf{X_{\widehat{f}}} := \big\{ \mathbf{x} \in \mathbf{G} \big| \widehat{f}(\mathbf{x}) \leq \mathbf{c}_{best} \big\}$.

The tree with vertices $\mathbf{V} \subseteq \mathbf{G}$  and edges $\mathbf{E} = {(\mathbf{v},\mathbf{w})}$ for vertices $\mathbf{v}\in \mathbf{V}$ and $\mathbf{w} \in \mathbf{V}$, is defined as $\tau := (\mathbf{V},\mathbf{E})$. From the current tree, the cost to come to a state $\mathbf{x} \in \mathbf{G}$ is given by the function $\mathbf{g_{\origtau}(x)}$. Any state that is not in the tree is assumed to have a cost $\infty$. If the optimal cost to come to a state is defined as $\mathbf{g(x)}$, then it is seen that, $\forall \mathbf{x} \in \mathbf{G}, \mathbf{\widehat{g}(x)} \leq \mathbf{g(x)} \leq \mathbf{g_{\origtau}(x)}$.

The cost of an edge between states $\mathbf{x},\mathbf{y} \in \mathbf{G}$ and the estimated cost of said edge are defined as $\mathbf{c(x,y)}$ and $\mathbf{\widehat{c}(x,y)}$. Any edge that intersects an obstacle is assumed to have an infinite cost thus defining that $\forall \mathbf{x},\mathbf{y} \in \mathbf{G}, \mathbf{\widehat{c}(x,y)} \leq \mathbf{c(x,y)} \leq \infty$. Calculating $\mathbf{c}(\cdot)$ is computationally expensive due to the need to account for obstacle collisions and dynamic constraints and therefore the heuristic estimate $\mathbf{\widehat{c}(\cdot)}$ is used to delay calculating this where possible. In the scenario where there are no dynamic constraints on the system, then an obstacle free edge $\mathbf{\widehat{c}(\cdot)} = \mathbf{c(\cdot)}$.

The Lebesque measure of a set is written as, $\lambda(\cdot)$, and the Lebesque measure of an n-dimensional unit ball, is $\zeta_n$. $|\cdot|$ refers to the cardinality of a set. $X\xleftarrow{+}\{x\}$ and $X\xleftarrow{-}\{x\}$ are shorthand notation for $X=X\cup\{x\}$ and $X=X\setminus\{x\}$ respectively.

\subsection{Informed tree algorithms}
\label{sec:tree_alg}

\begin{figure}[!t]
\removelatexerror
\begin{algorithm}[H]
    \label{alg:treeExpansion}
    \SetAlgoLined
    \caption{Informed tree expansion ($N,\mathbf{x_k},\theta_x^{(k)}$)}
    $\mathbf{G} \gets \mathbf{sample}(N)$\\
    $\mathbf{x}_{goal} \gets \mathbb{E}(\theta_{x}^{(k)})$\\
    $\mathbf{X}_{goal} \gets \big\{ \arg \displaystyle \min_{X \subseteq G}  \| \mathbf{x}_{goal}-\mathbf{X} \|_2 \mkern9mu \big| \mkern9mu |\mathbf{X}|=k_n\big\}$\\
    $\mathbf{V} \gets \mathbf{x}_k; \mkern9mu \mathbf{E} \gets \emptyset; \mkern9mu \mathbf{Q}_e \gets \emptyset; \mkern9mu \mathbf{Q}_v \gets \mathbf{V}$\\
    $r \gets 2\kappa\big(1+\frac{1}{n}\big)^{\frac{1}{n}}\big(\frac{\lambda(X_{\widehat{f}})}{\zeta_n}\big)^{\frac{1}{n}}\big(\frac{\log(|\mathbf{G}|)}{|\mathbf{G}|}\big)^{\frac{1}{n}}$ \\
    
    \While{$\mathbf{Q}_v \neq \emptyset$ \& $ \mathbf{g_{\origtau}(x)} \in \mathbf{X}_{goal} = \infty $}{
        $\mathbf{v}_m \gets \arg \displaystyle \min_{x \in Q_v} \mathbf{\widehat{f}(x)}$ \\
        $\mathbf{Q}_v \xleftarrow{-} \mathbf{v}_m$ \\
        $\mathbf{V}_{near} \gets \big\{ \mathbf{w} \in \mathbf{G} \mkern9mu  \big| \mkern9mu \| \mathbf{v}_m-\mathbf{w} \|_2 \leq r\big\}$ \\
        $\mathbf{Q}_e \xleftarrow{+} \big\{ (\mathbf{v}_m,\mathbf{w}) \in \mathbf{V}_{near} \big\}$\\
        \While{$\mathbf{Q}_e \neq \emptyset$}{
            $\mathbf{w}_m \gets \arg \displaystyle \min_{w \in Q_e} \mathbf{\widehat{c}(v_m,w)} + \mathbf{\widehat{h}(w)}$ \\
            $\mathbf{Q}_e \xleftarrow{-} (\mathbf{v}_m, \mathbf{w}_m)$\\
            \If{$\mathbf{g_{\origtau}(v_m)} + \mathbf{c(v_m,w_m)} < \mathbf{g_{\origtau}(w_m)}$}{
                \uIf{$\mathbf{w}_m \in \mathbf{V}$}{
                    $\mathbf{E} \xleftarrow{-} \big\{(\mathbf{v},\mathbf{w}_m) \in \mathbf{E} \big\}$}  
                \Else{
                    $\mathbf{V} \xleftarrow{+} \mathbf{w}_m$ \\
                    $\mathbf{Q}_v \xleftarrow{+} \mathbf{w}_m$
                }
                $\mathbf{E} \xleftarrow{+} (\mathbf{v}_m,\mathbf{w}_m)$\\
            }
        }
    }
   \KwRet $\mkern9mu \tau (\mathbf{V},\mathbf{E})$

    \end{algorithm}

\end{figure}

Algorithm \ref{alg:treeExpansion} outlines the tree expansion procedure during a single query event given the sampled states $\mathbf{G}$, the robots current location $\mathbf{x}_k$, and the current PDF of the source location $\theta_x^{(k)}$. 

\subsubsection{Initialisation (Alg \ref{alg:treeExpansion}, Lines 1:5)} 
To initialise, the goal state $\mathbf{x}_{goal}$ is defined as $\mathbb{E}(\theta_{x}^{(k)})$, from which the goal set $\mathbf{X}_{goal}$ is defined as the $k_n$ nearest $\mathbf{x} \in \mathbf{G}$. The tree vertices set $\mathbf{V}$ is set to $\mathbf{x}_k$,  the tree edges $\mathbf{E}$ and edge queue $\mathbf{Q}_e$ are set to empty, and the vertex queue $\mathbf{Q}_v$, is set to $\mathbf{V}$. The edge queues exist to track which vertex and edge should be processed for adding to the tree. The radius $r$ can also be defined during initialisation since only a single batch is being performed. The radius is calculated using the scaling parameter $\kappa$, the problem dimensionality $n$ and the number of sampled states $|\mathbf{G}|$, as described in \citep{Gammell2015}.

\subsubsection{Tree expansion (Alg \ref{alg:treeExpansion}, Lines 5:25)}
The tree is expanded until $\mathbf{Q}_v$ is empty (i.e. all $\mathbf{x} \in \mathbf{G}$ have been checked), or one of the states $\mathbf{x} \in \mathbf{X}_{goal}$ have a valid path in the tree. To determine which node should be selected for expansion, $\mathbf{v}_m$, the state $\mathbf{x} \in \mathbf{Q}_v$ with the lowest estimated cost from the start to the goal, $\widehat{\mathbf{f}}(\mathbf{x})$, is selected for expansion. The set $\mathbf{V}_{near}$ is defined around the expansion node which contains all states $\mathbf{w}$ in $\mathbf{G}$ that are within the radius $r$ of $\mathbf{v}_m$. The edges that connect all $\mathbf{w}$ to $\mathbf{v}_m$ are then added to the queue $\mathbf{Q}_e$.

Once the edge queue $\mathbf{Q}_e$ has been defined, each edge is processed by selecting the $\mathbf{w} \in \mathbf{Q}_e$ with the lowest estimated cost from the expanded vertex $\mathbf{v}_m$ to the goal, that passes through $\mathbf{w}$ (Alg \ref{alg:treeExpansion}, Line 13). The chosen edge is then removed from $\mathbf{Q}_e$. The edge will be added to the tree subject to the condition in Alg \ref{alg:treeExpansion}, Line 15. This is that the actual cost (including collisions) to $\mathbf{w}_m$ via $\mathbf{v}_m$ is less than the tree cost $\mathbf{g_\origtau}(\mathbf{w}_m)$ (if $\mathbf{w}_m$ is not already in the tree then this is guaranteed).

If $\mathbf{w}_m$ is already in $\mathbf{V}$, then its existing edge $(\mathbf{v},\mathbf{w}_m)$ is removed from $\mathbf{E}$ and the new edge $(\mathbf{v}_m,\mathbf{w}_m)$ is added. If $\mathbf{w}_m$ is not in the tree then it is added to $\mathbf{V}$ and also to the vertex queue $\mathbf{Q}_v$ (for future vertex expansion). The vertex expansion process is then ended when all edges of $\mathbf{v}_m$ have been checked i.e. $\mathbf{Q}_e$ is empty. When all vertices have been expanded or a path to the goal has been found, then the tree $\mathbf{\tau(V,E)}$ is returned for candidate selection (as shown in Fig \ref{fig:downsample}b, \ref{fig:addsample}b).

\begin{figure}[!t]
\removelatexerror
\begin{algorithm}[H]
    \label{alg:sampleAq}
    \SetAlgoLined
    \caption{Sample set acquisition ($\mathbf{\tau(V,E)}$)}
    $\mathbf{V} \xleftarrow{-} \big\{\mathbf{v} \in \mathbf{V} \mkern9mu | \mkern9mu \widehat{\mathbf{f}}(\mathbf{v}) > \mathbf{c}_{best} \big\}$\\
    $\mathbf{E} \xleftarrow{-} \big\{\mathbf{(v,w)} \in \mathbf{E} \mkern9mu | \mkern9mu \widehat{\mathbf{f}}(\mathbf{v}) > \mathbf{c}_{best} \mkern9mu \text{or} \mkern9mu \widehat{\mathbf{f}}(\mathbf{w}) > \mathbf{c}_{best} \big\}$\\

    \uIf{$|\mathbf{V}|>|\Sigma|$}{
         $r \gets 2\kappa\big(1+\frac{1}{n}\big)^{\frac{1}{n}}\big(\frac{\lambda(X_{\widehat{f}})}{\zeta_n}\big)^{\frac{1}{n}}\big(\frac{\log(|\mathbf{V}|)}{|\mathbf{V}|}\big)^{\frac{1}{n}}$ \\
        \While{$|\mathbf{V}|<|\Sigma|$}{
            $\mathbf{x}_m \gets \mathbf{sample}(1) \mkern9mu | \mkern9mu \widehat{\mathbf{f}}(\mathbf{x}_m) \leq \mathbf{c}_{best}+2\sigma_{\theta_x^{(k)}}$ \\
            
            $\mathbf{V}_{near} \gets \big\{ \mathbf{v} \in \mathbf{V} \mkern9mu  \big| \mkern9mu \| \mathbf{x}_m-\mathbf{v} \|_2 \leq r\big\}$ \\
            $\mathbf{Q}_e \xleftarrow{+} \big\{ (\mathbf{v},\mathbf{x}_m) \in \mathbf{V}_{near} \big\}$\\
            \While{$\mathbf{Q}_e \neq \emptyset$}{
                $\mathbf{v}_m \gets \arg \displaystyle \min_{v \in Q_e}  \| \mathbf{x}_m-\mathbf{v} \|_2$ \\
                $\mathbf{Q}_e \xleftarrow{-} (\mathbf{v}_m, \mathbf{x}_m)$\\
                \If{$\mathbf{g_{\origtau}(v_m)} + \mathbf{c(x_m,v_m)} < \mathbf{g_{\origtau}(x_m)}$}{
                    \uIf{$\mathbf{x}_m \in \mathbf{V}$}{
                        $\mathbf{E} \xleftarrow{-} \big\{(\mathbf{v},\mathbf{x}_m) \in \mathbf{E} \big\}$}  
                    \Else{
                        $\mathbf{V} \xleftarrow{+} \mathbf{x}_m$ \\
                    }
                    $\mathbf{E} \xleftarrow{+} (\mathbf{v}_m,\mathbf{x}_m)$\\
                }
            }
        }
    }
    \Else{
        $\mathbf{V}_m \gets \big\{ \arg \displaystyle \min_{V_m \subseteq V} \mathbf{g}(\mathbf{V}_m) \mkern9mu \big| \mkern9mu |\mathbf{V}_m|=|\Sigma|\big\}$\\  
        $\mathbf{V} \gets \mathbf{V}_m$\\
        $\mathbf{E} \xleftarrow{-} \big\{ \mathbf{(v,w)} \in E \mkern9mu | \mkern9mu \mathbf{v} \notin \mathbf{V} \mkern9mu \text{or} \mkern9mu \mathbf{w} \notin \mathbf{V} \big\}$
    }    
    \KwRet $\mkern9mu \tau (\mathbf{V},\mathbf{E})$

    \end{algorithm}

\end{figure}

\subsubsection{Sample set acquisition (Alg \ref{alg:sampleAq})}
\label{sec:tree_alg_acq}

The full tree $\tau$ from the initial search is groomed in Alg \ref{alg:sampleAq} in order to meet computational requirements of the \ac{IPP}. Firstly, the tree is pruned so that only vertices in $\mathbf{V}$ and edges in $\mathbf{E}$ which can possibly improve (or are part of) the current solution are kept for consideration (Alg \ref{alg:sampleAq}, Lines 1:2). This defines the ellipse of Fig \ref{fig:downsample}c, \ref{fig:addsample}c. In the case where no solution is found, i.e. $\mathbf{c}_{best}=\infty$, then $\mathbf{V}$ and $\mathbf{E}$ remain unchanged. 

After pruning, tree blossoming or tree culling will occur. The tree will be further expanded (blossoming, Fig \ref{fig:addsample}d) if there are not enough vertices in $\mathbf{V}$ to match $|\Sigma|$ (Alg \ref{alg:sampleAq}, Lines 4:21), or further reduced (culling, Fig \ref{fig:downsample}d) if there are more vertices in $\mathbf{V}$ than can be efficiently computed for their utility (Alg \ref{alg:sampleAq}, Line 23:25). 

For a given tree $\tau$ where $|\mathbf{V}|<|\Sigma|$, the tree is expanded with a new search radius $r$ defined by the current state of the tree (Alg \ref{alg:sampleAq}, Line 4). To give more potential sampling locations from which to evaluate utility, a new query state $\mathbf{x}_m$ is sampled from weighted free space $\mathbf{x} \in \mathbf{X}_{free}  \mkern9mu |  \mkern9mu \mathbf{D}_{\phi}$ subject to the constraint that $\widehat{\mathbf{f}}(\mathbf{x}) \leq \mathbf{c}_{best}+2\sigma_{\theta_x^{(k)}}$ (Alg \ref{alg:sampleAq}, Line 6). This defines the ellipse of Fig \ref{fig:addsample}d. The addition of including source location uncertainty is necessary in order to account for the fact that it is probabilistically likely that the true $\mathbf{x}_s$ lies within 2 standard deviations of the goal $\mathbf{x}_{goal}$ and therefore new exploratory samples should be drawn accounting for this. Furthermore, as $\mathbf{c}_{best} \rightarrow 0$, $\mathbf{x} \in \mathbf{X}_{free} \rightarrow \emptyset$ and therefore to avoid this local minimum, the robot should attempt to search in an area relative to the uncertainty of its estimate. Alg \ref{alg:sampleAq}, Lines 7:20 follows closely to the tree expansion process of Alg \ref{alg:treeExpansion}, Lines 11:23. The difference being that we are now attempting to connect the unconnected state $\mathbf{x}_m$ to one of the vertices in the tree (as opposed to expanding the tree into the unconnected set). If a sampled state $\mathbf{x}_m$ cannot be connected to the tree, then the criterion at Alg \ref{alg:sampleAq}, Line 12 will always fail and a new state will be initiated with $\mathbf{V}$ and $\mathbf{E}$ remaining unchanged. Once the number of vertices in the tree equals $|\Sigma|$, then the final state of $\tau$ is returned for utility calculation. This expansion process is similar in formulation to Alg \ref{alg:treeExpansion} however single query states are attached to the tree set as opposed to the tree expanding into the unconnected set.   

If for a given tree $\tau$, where $|\mathbf{V}|\geq |\Sigma|$, the tree is culled as per Alg \ref{alg:sampleAq}, Lines 23:25. A query set of vertices $\mathbf{V}_m$ is defined which takes the first $|\Sigma|$ number of $\mathbf{v} \in \mathbf{V}$ which have the lowest tree cost. Since $\mathbf{V}$ has already been pruned, the remaining branches in $\mathbf{V}_m$ are towards the goal whilst allowing for some exploratory states around the path to the goal. $\mathbf{V}$ is then updated to the new set $\mathbf{V}_m$ and any edges that contain old vertices are removed from $\mathbf{E}$ (Alg \ref{alg:sampleAq}, Line 25). Given the scenario where $\mathbf{c}_{best}=\infty$, then $\mathbf{V}_m$ will be a non-directional set of nodes that were connected to $\mathbf{x}_k$ during tree expansion.  

\section{IPP Utility calculation}
\label{sec:utility}
Given the robot has acquired a set candidate trajectories $\Sigma$, the robot needs to select the trajectory and the sample location that will minimise Eq. \eqref{eqn:IPP}. Several definitions of the utility function $\Psi(\cdot)$ can be used together with parametric modelling techniques such as the Bayesian inference used here. Based on the results attained in \citep{Hutchinson2018}, the Entrotaxis measure of information gain has proven to be effective in source search and therefore is the chosen metric when defining the utility function. Entrotaxis attempts to find the most informative location by considering the entropy of the predictive measurement distribution at a sample location, therefore, we define this location as $\sigma(1)$ given the start of the trajectory $\sigma(0)$. In Entrotaxis, the Shannon Entropy $\mathbf{H}(\cdot)$ is used as the expected information measure as follows:

\begin{multline}
    \Psi(\sigma)=-\int P(\widehat{\mathbf{z}}_{k+1}(\sigma)|\mathbf{z}_{1:k}) 
    \log P(\widehat{\mathbf{z}}_{k+1}(\sigma)|\mathbf{z}_{1:k})d\widehat{\mathbf{z}}_{k+1}
    \label{eqn:Ent}
\end{multline}

where $\widehat{\mathbf{z}}_{k+1}(\sigma)$ refers to the unknown measurement at the potential sampling position of $\sigma(1) \in \Sigma$. This unknown measurement will not be known until the location is physically sampled and therefore the probability of the expected number of particle encounters $P(\widehat{\mathbf{z}}_{k+1}(\sigma)|\mathbf{z}_{1:k})$ is derived using the posterior distribution of the source $\Theta_{k}$:
\begin{equation}
\begin{split}
    &P(\widehat{\mathbf{z}}_{k+1}(\sigma)|\mathbf{z}_{1:k}) \\
    &= \int_{\Theta_{k+1}} P(\widehat{\mathbf{z}}_{k+1}(\sigma),\Theta_{k+1}|\mathbf{z}_{1:k})d\Theta_{k+1} \\
    &=\int_{\Theta_{k+1}}P(\widehat{\mathbf{z}}_{k+1}(\sigma)|\Theta_{k+1})P(\Theta_{k+1}|\mathbf{z}_{1:k})d\Theta_{k+1} \\
    &\approx \sum_{i=1}^{n} w_k^{(i)} \cdot P(\widehat{\mathbf{z}}_{k+1}(\sigma)|\Theta_{k+1}^{(i)})
\end{split}    
    \label{eqn:zk+1}
\end{equation}
where the weighted samples $\{\Theta_{k}^{(i)}, w_{k}^{(i)}\}_{i=1}^{n}$ constitutes the posterior distribution $P(\Theta_{k}|\mathbf{z}_{1:k})$ and $\Theta_{k+1}^{(i)}$ = $\Theta_{k}^{(i)}$ \citep{Hutchinson2019b}. To reduce computational load, a much smaller number of samples $\{\Theta_k^{(l)},1/n_z\}_{i=l}^{n_z}$ is resampled from the full posterior, where $n_z \ll n$, to give a possible future measurement set of $\{\widehat{\mathbf{z}}_{k+1}^{(l)}\}_{l=1}^{n_z}$ and reduce Eq. (\ref{eqn:zk+1}) to:

\begin{align}
    P(\widehat{\mathbf{z}}_{k+1}(\sigma)|\mathbf{z}_{1:k}) \approx \frac{1}{n_z}\sum_{l=1}^{n_z} \delta\big(\widehat{\mathbf{z}}_{k+1}-\widehat{\mathbf{z}}_{k+1}^{(l)}\big)
        \label{eqn:zk+1_2}
\end{align}

Substituting Eq. \eqref{eqn:zk+1_2} into Eq. \eqref{eqn:Ent} allows the entropy for performing the trajectory $\Psi(\sigma)$, to be approximated by summing over the possible future measurements. 
\begin{align}
    \Psi(\sigma) \approx \frac{1}{n_z} \sum_{l=1}^{n_z} \widehat{w}_{k+1}^{(i,l)} \log \widehat{w}_{k+1}^{(i,l)}
\end{align}
$\Psi(\cdot)$ is then calculated for all $\sigma \in \Sigma$ and minimised as per Eq. \eqref{eqn:IPP}, to give the optimal trajectory $\sigma^*_k$. The Entrotaxis reward function is calculated every time a new set of $\Sigma$ is defined from the informed tree search. 

{Whilst the Entrotaxis utility function has been utilised, any information theoretic measure may be used in its place. The exploratory effects of such measures ensure that the searching agent does not get stuck in minima around the goal set and will tend the agent to continue picking samples that minimise the estimation uncertainty (encouraged by the tree blossoming effect). This feature also helps in the case of a misleading prior, where there is a mismatch between the $\mathbf{s}$ and $\mathbb{E}(\mathbf{s})$, since the Entrotaxis utility function will pick samples that update the source estimate away from the incorrect region.}

At this point in the system, the \ac{IPP} has taken an informed set of potential trajectories from the informed tree search, performed predictive modelling on this subset of samples to calculate predicted information gain for each trajectory, and then chosen the optimal solution to be executed by a low-level path planner. 

This defines a single control loop of the proposed source search system. The inference, informed tree search and utility calculation procedures are then repeated iteratively as per Fig. \ref{fig:systemOverview} until an end constraint on the system is met, e.g., a time budget. 

\section{Simulation} \label{sec:simu}
\begin{figure*}[!t]
    \centering
    \includegraphics[width=\textwidth]{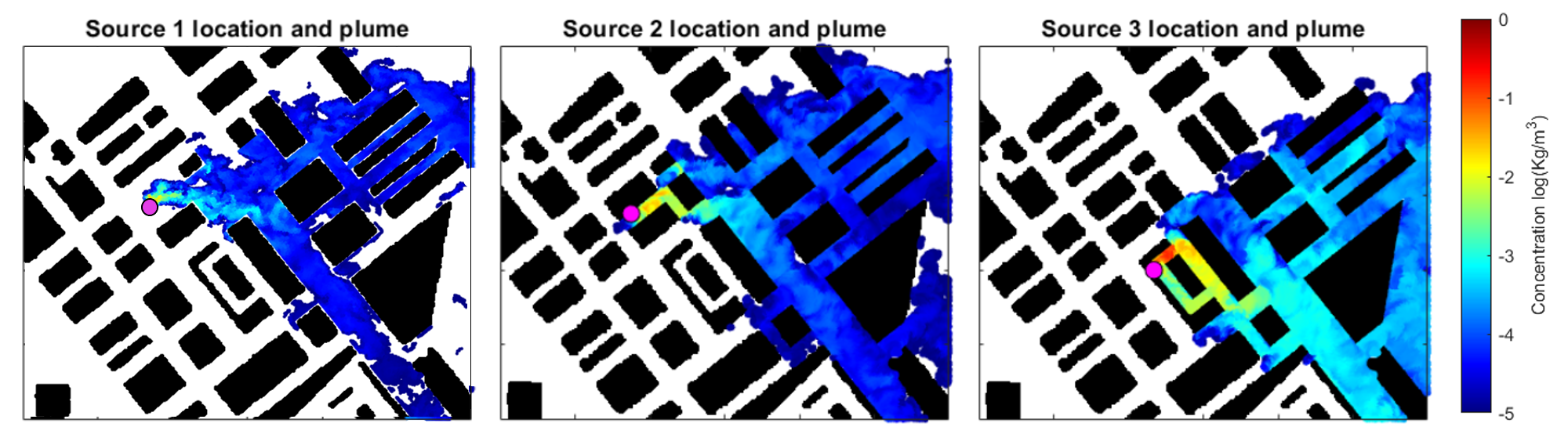}
    \caption{Snapshot concentration colour map for each of the 3 tested sources, superimposed onto the occupancy map (time$=0$s). Concentrations are coloured as $\log(\text{Kg/m}^3)$ with concentrations below $10\mu \text{g/m}^3$ not coloured (source 3's plume has been artificially inflated by $10^3$ for visual aid). Source location is shown as a magenta circle. Sources 1 \& 2 are vented sources with a perturbation of 1m/s and source 3 is a passive source with no excitation.}
    \label{fig:sources}
\end{figure*}

A set of simulation studies have been carried out to verify the proposed autonomous search algorithm in a large scale, outdoor, feature rich environment. The staging for the study is the DAPPLE dispersion scenario \citep{Martin2010}, which is generated by an experimentally validated CFD simulation of a source release under steady wind conditions in urban London. This dataset contains a complex series of urban canyons causing local wind field instabilities, which is not only challenging for the algorithms to predict the source, but also suitable to test the efficient path finding ability of the main contribution of this paper, i.e. the informed tree search.

There are 3 source locations within the same DAPPLE domain that can be tested as shown in Fig. \ref{fig:sources}. Sources 1 \& 2 are modelled as active vent releases of outlet velocity $1$m/s whilst source 2 is a passive release where the main method of transportation is the local wind field formed inside the urban environment. Sources 2 \& 3 are modelled as ground releases whilst source 1 is a release on top of a building. The corresponding occupancy grid for source 1 is less dense due to being at a higher altitude than some of the surrounding buildings, and due to this, the corresponding sample distribution is also slightly different. At the initialisation of the each source i.e. $k=0$s, the plume structure is quasi-stable and therefore is only locally fluctuating, whilst the main plume structures are stable. All sources are modelled as constant release, matching the assumption made in the estimation model, however each source configuration matches the estimation model with varying accuracy as will be shown in the results.  

The popular Entrotaxis approach is used as the benchmark in this paper from which conclusions about the proposed algorithm will be drawn. To ensure functionality in a feature rich environment, Entrotaxis is slightly modified for simple obstacle avoidance, so that any candidate trajectories $\sigma \subseteq \mathbf{X}_{\mathrm{obs}}$ are discarded before utility calculation.

In addition to the benchmark algorithm, Entrotaxis-Jump \citep{Zhao2020} is also compared in the study. Similar to the proposed algorithm, Entrotaxis-Jump is based on the Entrotaxis utility function and has also been designed for use in dense urban environments. Entrotaxis-Jump is originally presented with four deterministic path planning actions of $\{\uparrow, \rightarrow, \downarrow, \leftarrow\}$, however we extend the algorithm to match the planning horizon of standard Entrotaxis in Table \ref{tab:IPPParam} (detailed implementation can be found in Appendix \ref{app:jump}).

This paper presents two unique additions to the source search process: informed tree search and sample generation. As discussed previously, the sample generation technique is proposed to minimise the effect of buildings in the sampling process and as such, it is expected that the sample generation technique will have the largest improvement on sources that do not well match the \ac{IP} plume of the estimation model. To prove the efficacy of the sample generation methodology, the system is also tested without this feature (termed as uniform tree search) and compared alongside the fully informed tree search algorithm.

\subsection{Test Setup}
\begin{figure*}[!t]
    \centering
    \includegraphics[width=0.95\textwidth]{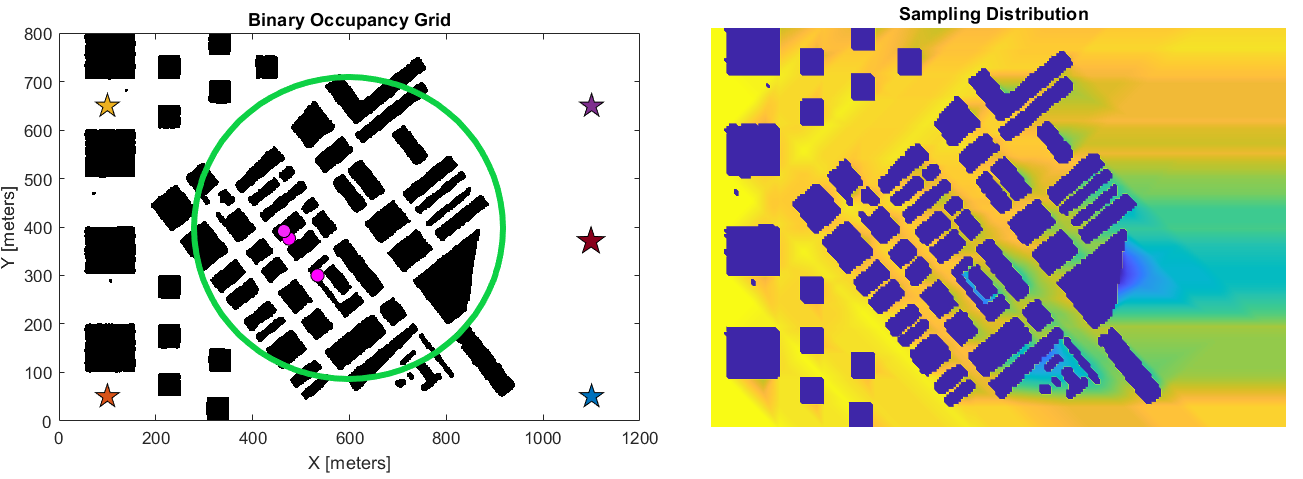}
    \caption{Left: 960,000m$^2$ search area showing obstacles (black), 5 starting positions (coloured stars), 3 source locations (magenta circles) and the initial prior source location area defined in $\theta_{x,y}$ (green circle). Right: Sample generation distribution attained from Dijkstra's search with $\mathbf{X}_{inlet}=[0,0:800]$ (for sources 2 \& 3). Yellow colouring is a high likelihood whilst blue colouring is a low likelihood.}
    \label{fig:sim+dist}
\end{figure*}

Each control strategy is tested across all 3 sources with the same prior parameters of $\Theta$ initiated for each source, as shown in Table \ref{tab:simParam}. {Prior distributions are implemented following literature examples as typical starting source search conditions \citep{Hutchinson2019a,Ristic2017}. Gaussian distributions are set on the source location to implement the domain knowledge that the source is most likely to be at the centre of the search area (uniform distributions may also be used given no user domain information).} Testing three sources ensures that results can be evaluated for their robustness in differing configurations, as opposed to the circumstance that a particular method favours a single type of source release. 

To further test adaptability of the proposed algorithm, 5 starting locations are also tested for each source (shown in Fig. \ref{fig:sim+dist}). Locations 1-5 are situated at [1100,50], [1100, 325], [1100,650], [100,50] \& [100,650] respectively. Locations 1-3 are downwind of the plume ({aligned along the x-axis}) and represent typical favourable starting locations for source search. These downwind locations are used to determine the general performance of each control strategy. Despite the fact that source search is typically started downwind, to test the robustness of the algorithms, two further upwind locations have also been tested that represent unfavourable starting locations.

Model parameters for the estimation engine are shown in Table \ref{tab:simParam} alongside the ground truth values of the sources. Sources 1, 2, \& 3 locations are $[466,392]$, $[475,376]$ \& \\
$[534,300]$ respectively.  

\begin{table}[!t]
    \centering
    \caption{Model parameters for the estimation engine prior parameters alongside ground truth values.}
    \label{tab:simParam}
    \begin{tabular}{lll}
        Model Parameters       & Ground Truth  & Initial Prior \\
        \hline
        \hline
       % Dispersion Model& GP                & GP \\
        $x$ position $x_s$              & source$_{1,2,3}$    & $\mathcal{N}(600,100)$ \\
        $y$ position $y_s$              & source$_{1,2,3}$    & $\mathcal{N}(400,100)$ \\
        $z$ position $z_s$              & $13.6,0,0$ m      & $\mathcal{N}(1,0.5)$ \\
        Release rate $q_s$              & {$1.11, 1.14, 1$kg/s}   & $\gamma(2,1)$ \\
        Wind speed $u_s$              & $2.5$m/s    & $\mathcal{N}(2.5,2)$ \\
        Wind direction $\phi_s$           & $270^o$    & $\mathcal{N}(270,10)$ \\
        Diffusivity $d_s$   & $-$       & $\mathcal{N}(1,2)$ \\
        Particle lifetime $\origtau_s$   & $-$       & $\mathcal{N}(8,2)$ \\
        PF Particles $n$       & $-$         & $20,000$ \\
        Effective ratio $\eta$               & $-$         & 0.5 \\
        \hline
    \end{tabular}
\end{table}

Key parameters for the Entrotaxis, Entrotaxis-Jump and the Informed tree search are outlined in Table \ref{tab:IPPParam}. To ensure a fair comparison, all three control methods are subject to the same number of utility calculations per step (equating to $|\Sigma|$), and the predictive measurements of the particle filter $n_z$ also remains constant. From the obstacle map of the DAPPLE domain, the sample generation distribution for the informed tree search is shown in Fig. \ref{fig:sim+dist}.

\begin{table}[!t]  
\caption{Parameters for \ac{IPP}}
    \label{tab:IPPParam} 
    \centering
       \begin{tabular}{ll}
        Operational Parameters  & Robot   \\
        \hline
        \hline
        Number of nodes, $N$            & 4000 \\
        Goal set neighbours, $k_n$            & 5 \\
        Sample evaluations, $|\Sigma|$         & 16 \\
        Entrotaxis/Jump step size                           & [10,20]m\\
        Entrotaxis/Jump step directions                     & $[0^o,45^o, \dots ,315^o]$ \\
        Predictive measurements, $n_z$    & 40 \\
        \hline
    \end{tabular}
\end{table}

To model the sampling robot, we assume a mobile sensor (e.g. unmanned ground/aerial vehicle) with parameters outlined in Table \ref{tab:UAVParam}. The vehicle is fitted with a single fast response chemical sensor capable of detecting down to a minimum concentration of 0.1g/m$^3$. Due to the limitations of the dataset, concentration data is only available at a sampling height of 5m for sources 2 \& 3, and 15m for source 1, and therefore it is assumed that the robot has a fixed sampling height. The sensing robot is assumed to travel at a constant velocity. The time budget allowed is more conducive to a ground vehicle however, conclusions for suitability on a UAV can still be drawn by analysing the presented results at a lower time budget.  

\begin{table}[!t]  
\caption{Operational parameters for simulated robot}
    \label{tab:UAVParam} 
    \centering
       \begin{tabular}{ll}
        Operational Parameters  & Robot   \\
        \hline
        \hline
        Motion Model            & constant velocity \\
        Velocity                & 2m/s \\
        Time budget           & 3600s\\
        Sensor threshold    & 10mg/m$^3$\\
        Sampling altitude     & 15,5,5 m\\ 
    \end{tabular}
\end{table}

\subsection{Results} \label{sec:results}
To analyse the performance of a searching strategy, the weighted \ac{RMSE} between the source location $\mathbf{s}_{x,y}$ and the current posterior estimate of the Bayesian inference $\theta_{x}^{(k)}$ is recorded after each sampling event. The equation for calculating the weighted \ac{RMSE} is:
\begin{align}
    RMSE_k=\sqrt{\sum_{i=1}^n\mathbf{w}_k^{(i)}\|\theta_{x}^{(k)}-\mathbf{s}\|_2^2}
\end{align}

The \ac{SR} and \ac{MST} are also studied, since they are common criteria in evaluating source search algorithms. A source is defined as successfully resolved when the \ac{RMSE} of the inference engine drops below $50$m during a single test run. The \ac{MST} is defined as the time taken for the source to be resolved, averaged across all successful runs. As such, \ac{MST} is used to determine how efficient the searching process is, whilst \ac{SR} determines the reliability of the search.

\subsubsection{Individual source analysis}

\begin{figure*}[!t]
    \centering
    \includegraphics[width=0.98\textwidth]{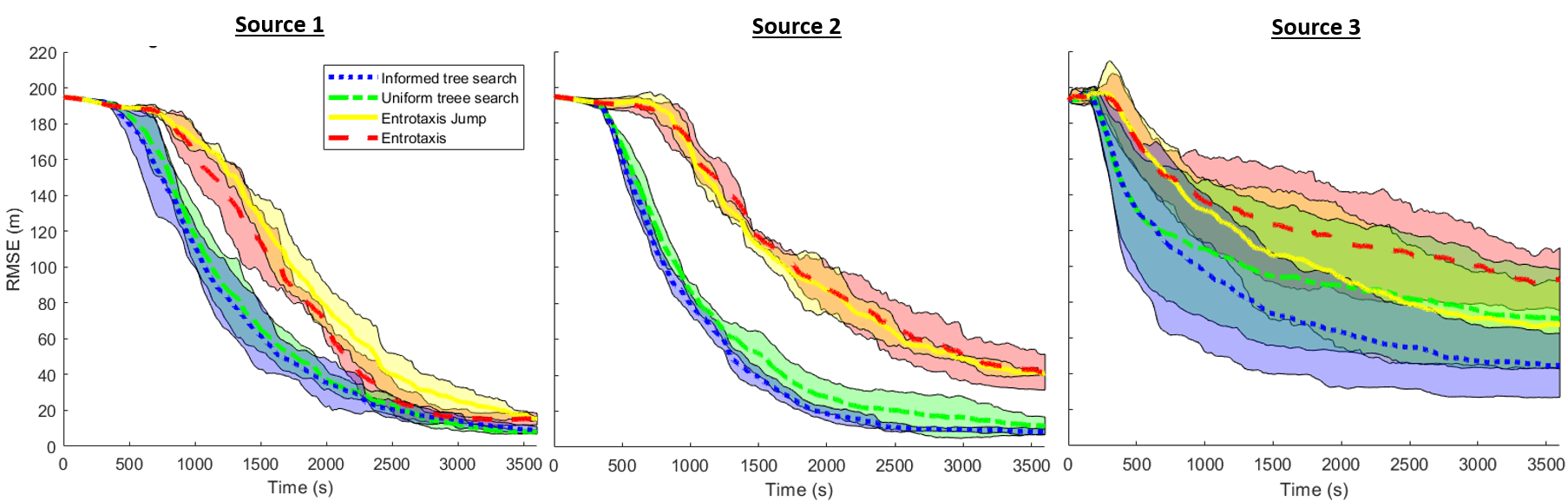}
    \caption{RMSE over time averaged across all 360 Monte Carlo simulations for sources 1, 2 \& 3 for the downwind start locations. One standard deviation bounds from the mean are shown in the corresponding mean line colour.}
    \label{fig:downwind_RMSE}
\end{figure*}

For the 3 different sources with the downwind starting locations, Fig. \ref{fig:downwind_RMSE} shows the average \ac{RMSE} at each time step, as well as $1\sigma$ upper and lower bounds of the 360 Monte Carlo runs per method. Downwind starting locations are only selected for the individual source analysis to better show the convergence performance without accounting for the algorithms' robustness to unfavourable start conditions. Table \ref{tab:downwind} also shows the \ac{SR} and \ac{MST} of the same simulations. It can be seen that for all sources, the searching efficiency of the proposed tree search has improved significantly over Entrotaxis and Jump (shown by a lower \ac{MST} and faster \ac{RMSE} reduction rate). Search reliability has also increased as shown by the \ac{SR} of the tree searches. 

\begin{table}[!h]
    \centering
\caption{Downwind location SR and MST for each of the three sources} 
    \label{tab:downwind}
       \begin{tabular}{lccccc}
       \multicolumn{2}{c}{Source} & \multicolumn{4}{c}{Algorithm} \\
       \quad & \quad & Entro & Jump & Uniform & Informed \\
        \hline
        \hline
        % $\mathbf{s_1}$  & 84\% (950) & 86\% (1119) \\
        % \hline
        $\mathbf{s_1}$& SR  & 96\% & 96\% & \textbf{100}\% & \textbf{100}\%\\
        \quad & MST & (1823) & (1939) & (1460) & (\textbf{1367}) \\
        \hline
        $\mathbf{s_2}$& SR  & 74\% & 78\% & 97\% & \textbf{98}\% \\
        \quad & MST & (2082) & (1943) & (1333) & (\textbf{1157}) \\
        \hline
        $\mathbf{s_3}$& SR  & 42\%  & 58\%  & 59\%  & \textbf{78}\% \\
        \quad & MST & (1876) & (1821) & (1484) & (\textbf{1307}) \\
        %\hline
        % \textbf{all} & SR  & 76\%  & 76\%  & 85\%  & \textbf{92}\% \\
        % \quad & MST & (1820) & (1915) & (1417) & (\textbf{1275}) \\
    \end{tabular}

\end{table}

When comparing between the sources, each source's properties must be first considered. Of the 3 tested sources, source 1 \& 2 match the \ac{IP} model the most closely with source 3 being the least well modelled. This is due to the location of source 3 being most affected by the environment structure. Whilst this can be seen visually in Fig. \ref{fig:sources}, it is also shown in the results by studying the final \ac{RMSE} when examining the sources individually. For source 1 \& 2, the average converged \ac{RMSE} is $11$m \& $25$m, respectively, against an average of $67$m for source 3, proving the greater mismatch between the estimation model and the actual source. Source 1 is the easiest on the path planning front since, as explained previously, the occupancy map is more sparse as shown by the mutual convergence of all the planners. Source 2 is more challenging on the path planning perspective and the ability of the tree search to adaptively dictate sample steps based on the posterior variance is clearly shown by the disparity in converged \ac{RMSE}.   

In section \ref{sec:dist}, it is argued that the preferential sampling distribution should see a gain in efficiency when estimating sources that do not well match the model. Based on this notion, it is expected that the informed search should see a larger efficiency gain over the uniform search on source 3, whereas only a marginal performance gain is expected on source 2 and similar performance on source 1. This is clearly shown in Fig. \ref{fig:downwind_RMSE} as well as in the \ac{SR} and \ac{MST} values of Table \ref{tab:downwind}. This result shows that the informed searching method can help make the source search process more efficient (especially in complex dispersion scenarios) and allows further research into how other prior sampling distributions may be incorporated for more efficiency gains.

Analysing the performance of the Jump strategy, it can be seen that Jump performs well in the more difficult source 3 (comparable with the uniform search), but has a generally slower \ac{MST} due to jumps around buildings causing extended traversals between sampling events. For sources 1 \& 2, comparable performance with Entrotaxis is seen. Since one set of parameters are set across the sources to ensure fairness of comparison, the Jump algorithm does not necessarily perform optimally in each case and this highlights the need for an adaptive algorithm such as the proposed method. This is seen in source 1 where the plume near the source has significantly fewer obstacles. This causes Jump to unnecessarily jump around some buildings leading to a longer \ac{MST} despite the same \ac{SR} as Entrotaxis, indicating a need for a larger $n_{jump}$ value than in the other sources.

\subsubsection{Overall evaluation}
\begin{figure}[!t]
    \centering
    \includegraphics[width=0.9\columnwidth]{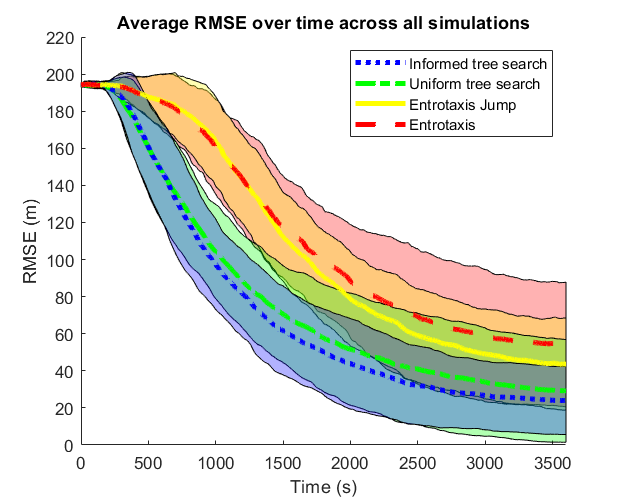}
    \caption{RMSE w.r.t. time, averaged across all 600 Monte Carlo runs for each method over three scenarios . 1 standard deviation bounds from the mean are shown in the corresponding mean line colour.}
    \label{fig:RMSE_all}
\end{figure}

For the overall evaluation, the 2 upwind start locations are included to demonstrate each algorithm's general performance including robustness to unfavourable start locations. Fig. \ref{fig:RMSE_all} shows the average \ac{RMSE} at each time step, as well as $1\sigma$ upper and lower bounds of the 600 Monte Carlo runs per method. It is clear to see that even with upwind locations, overall the informed tree search shows significant improvement over the standard Entrotaxis and Entrotaxis-Jump approaches. The informed search shows a much greater initial rate of error reduction and well as having a better average final \ac{RMSE} at $k=3600$s ($24$m vs $29$m vs $44$m vs $54$m). Furthermore, the variance of the proposed method is much narrower than the other methods, showing that performing the informed tree search gives more consistence. Table \ref{tab:allResults} shows the proposed search method also drastically increases the \ac{SR} as well as the \ac{MST} over both Entrotaxis and Entrotaxis-Jump.

The overall results also back the findings of \citet{Zhao2020}, by showing Entrotaxis-Jump outperforms the standard algorithm (in urban environments) with a significantly improved \ac{SR} despite a comparable \ac{MST}. Fig. \ref{fig:RMSE_all} shows how initially Jump and Entrotaxis have the same performance, due to the initial search not being in a cluttered region. This shows that Entrotaxis-jump will only outperform Entrotaxis given a dense urban region (its designed purpose), whereas our proposed method also outperforms Entrotaxis in open regions due to the tree search stemming towards the likely source location thus providing more potential sampling locations in an informed direction.
% It can also be seen that Jump performs more robustly than Entrotaxis in the upwind directions due to its ability to search quickly around buildings when there isn't much information to be gained in the immediate area.

\begin{table}[!h]
    \centering

\caption{\ac{SR} and \ac{MST} across all simulations} 
    \label{tab:allResults}
       \begin{tabular}{lccc}
       \multicolumn{4}{c}{Overall average SR (MST)} \\
       Entro & Jump & Uniform & Informed \\
        \hline
        \hline
        68\% (1813) & 76\% (1810) & 87\% (1394) & \textbf{91}\% (\textbf{1280})\\
    \end{tabular}
    
\end{table}

\begin{figure*}[!t]
    \centering
    \includegraphics[width=0.8\textwidth]{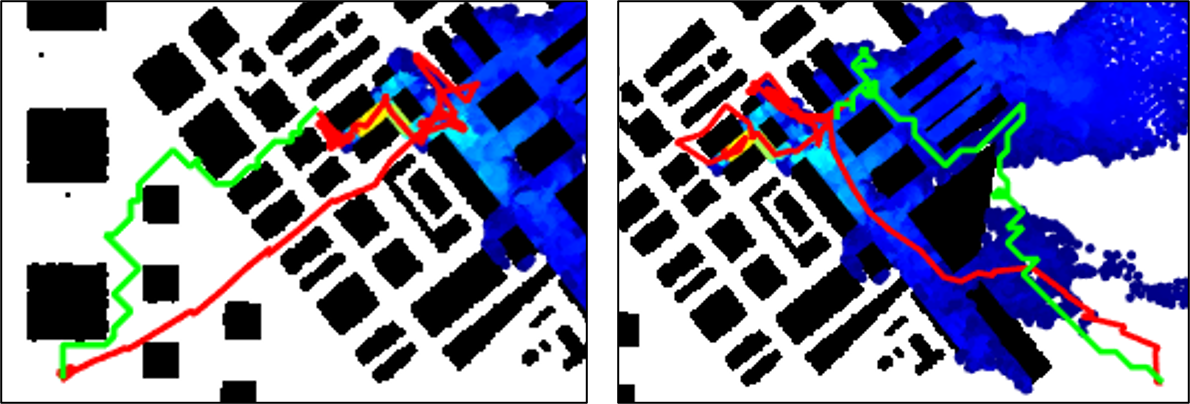}
    \caption{Example trajectories for estimating source 2 with a downwind and upwind starting position, from $t=0$ to $t=1200$s of simulation time. Red line indicates the informed tree historic trajectory and green line represents the Entrotaxis trajectory.}
    \label{fig:example}
\end{figure*}

For context, example trajectories for a single run of the informed tree search and Entrotaxis are shown in Fig. \ref{fig:example}. The efficiency bonus of using informed trees is clear to see as 1200s into the simulation, the informed tree approach has navigated the robot to the vicinity of the source location, whereas Entrotaxis has not reached the main plume.

% \begin{figure}[!t]
%     \centering
%     \includegraphics[width=\columnwidth]{boxplot2.png}
%     \caption{Boxplot of the RMSE at $k=3600$s for each starting location compared across control methods. Starting position numbers refer to locations in Fig. \ref{fig:sim+dist}.}
%     \label{fig:boxplot}
% \end{figure}

\section{Conclusion} \label{sec:conc}
In this paper, we integrate obstacle avoidance trajectory generation with a source term estimation engine to deliver autonomous search of an airborne release in urban environments. By combining a state-of-the-art parametric inference model alongside a single batch informed tree search algorithm, we are able to efficiently navigate a sampling robot in an informed manner towards a source location. Furthermore, local minima are avoided by taking exploratory actions relative to the uncertainty of the source location estimate. 

As demonstrated in the simulation studies, the presented approach is reliably capable of localising a source within a large-scale and feature-rich environment against a variety of source terms and varied start conditions. The informed tree search is shown to far outperform the baseline Entrotaxis and Entrotaxis-Jump approaches in an urban environment when evaluating the same number of possible future sampling locations, proving the efficiency bonus of planning trajectories towards the goal. The addition of an informed sample generation algorithm, that attempts to account for the fundamental mismatch between the plume model and the actual complex flow around obstacles, is also shown to provide a further efficiency gain with regards to the rate of error reduction that can be achieved. Since the batch tree search is capable of accepting any prior sampling distribution, future work can look at alternative distribution to make the search process even more efficient and robust.
Overall, the results demonstrate that the proposed framework is capable of guiding the sensing robot to respond to emergency \ac{CBRN} events in urban environments. 

\begin{acknowledgements}
This work is supported in part by UK \ac{DSTL} under the project No. 1000155749 and in part by the \ac{EPSRC} under the project No. 2126619. The author would like to thank Tim Foat at the \ac{DSTL} for providing the DAPPLE experiment dataset used as the ground truth in this paper.
\end{acknowledgements}


\bibliographystyle{spbasic}      % basic style, author-year citations

\bibliography{library}   % name your BibTeX data base
% % Non-BibTeX users please use
% \begin{thebibliography}{}
% %
% % and use \bibitem to create references. Consult the Instructions
% % for authors for reference list style.
% % Format for Journal Reference
% Author, Article title, Journal, Volume, page numbers (year)
% Author, Book title, page numbers. Publisher, place (year)
% \end{thebibliography}

\appendix

\section{Entrotaxis-Jump implementation} \label{app:jump}

In the original implementation of Entrotaxis-jump \citep{Zhao2020}, the number of possible future control actions, $\sigma_k^{k+1}$, is limited to the deterministic directions of $\uparrow, \rightarrow, \downarrow, \leftarrow$ with a step size that must be predetermined. In the original study, an area approximately $25\%$ of the area of that in this study is searched and a step size of $5$m is chosen. Another key difference between the two studies is that the ratio of $\mathbf{X}_{\mathrm{free}} : \mathbf{X}$ is much greater in our study, and therefore the number of times that the searcher will be hindered by an obstacle will be appreciably lower. Based on these two main differences, further considerations must be made for the implementation of Entrotaxis-Jump in this study.

To match the standard Entrotaxis implementation and provide fair comparison, the same 8 possible move directions (as detailed in Table \ref{tab:IPPParam}) and the same two step sizes are used for consideration in $\Sigma$. Due to incorporating two possible sampling locations per direction, at each move event there are two chances that the jump counter, $n_{\mathrm{jump}}$, is triggered per direction. Due to this fact and also the aforementioned statement of searching in a greater $\mathbf{X}_{\mathrm{free}} : \mathbf{X}$, the parameter values for both $n_{\mathrm{jump}}$ (jump counter) and $m_{\mathrm{jump}}$ (jump memory) must be investigated for our scenario.

\begin{figure}[ht]
    \centering
    \includegraphics[width=0.9\columnwidth]{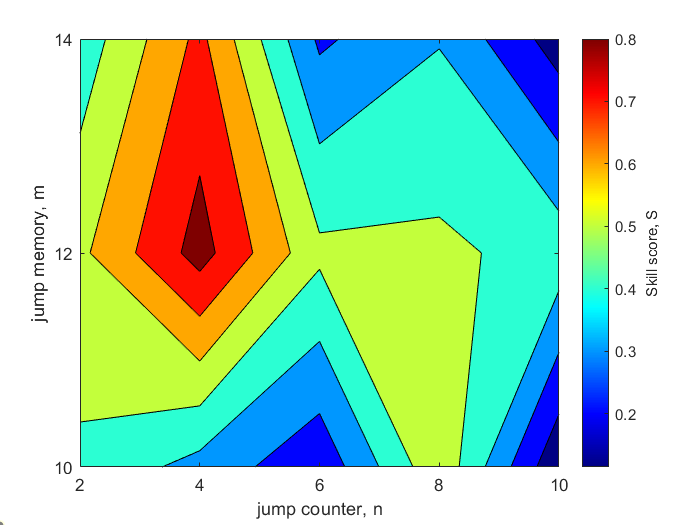}
    \caption{Skill score contour plot for the combinations of $m_{\mathrm{jump}}$ and $n_{\mathrm{jump}}$}
    \label{fig:jumpskill}
\end{figure}

To this end, the same $n_{\mathrm{jump}}$ \& $m_{\mathrm{jump}}$ sensitivity study as performed in the original work is employed on all sources ($\mathbf{s}_1$, $\mathbf{s}_2$, $\mathbf{s}_3$) with start location $[1100, 325]$, repeated 40 times each. Jump parameter values tested are $n_{\mathrm{jump}} = \{2, 4, 6, 8, 10\}$  and $m_{\mathrm{jump}} = \{10, 12, 14\}$. The skill score equation from the original work is used for evaluation.

\begin{align}
    S^i_{mst} &= (MST_{max} - MST^i) / (MST_{max} - SR_{min}) \\
    S^i_{sr} &= (SR^i - SR_{min}) / (SR_{max} - SR_{min}) \\
    S &= w_{sr} \cdot S_{sr} + w_{mst} \cdot S_{mst}
\end{align}
where  $w_{sr}=0.5$ and $w_{mst}=0.5$ as per default. \ac{SR} and \ac{MST} are calculated the same as in Section \ref{sec:results}. The skill score aggregated from all sources for all 40 repeats are shown in Fig. \ref{fig:jumpskill}. The results show that our scenario favours a lower jump threshold, $n_{\mathrm{jump}} = 4$, and the middle value $m_{\mathrm{jump}} = 10$. This can be primarily explained by having more free space thus the searcher is less likely to trigger a necessary jump within its memory of $m_{\mathrm{jump}}$. Therefore, these Entrotaxis-Jump parameters are set when running all the simulation configurations of our main study.

\end{document}